\documentclass[review]{elsarticle}

\usepackage{lineno,hyperref}
\modulolinenumbers[5]

\journal{Journal of \LaTeX\ Templates}

\usepackage{booktabs}
\usepackage{url}
\usepackage{multicol, blindtext}
\usepackage{multirow}
\usepackage{bbm}
\usepackage{amsmath}
\usepackage{amssymb}
\usepackage{pifont}
\usepackage{xcolor}
\usepackage{adjustbox}
\usepackage{pgfplots}
\newcommand{\cmark}{\ding{51}}%
\newcommand{\xmark}{\ding{55}}

\begin{document}

\begin{frontmatter}

\title{You Only Watch Once: A Unified CNN Architecture for Real-Time Spatiotemporal Action Localization}


\author{Okan K\"op\"ukl\"u\corref{mycorrespondingauthor}} 
\cortext[mycorrespondingauthor]{Corresponding author}
\ead{okan.kopuklu@tum.de}

\author{Xiangyu Wei\corref{}} 
\author{Gerhard Rigoll\corref{}}
\address{Institute for Human-Machine Communication, Technical Univ. of Munich, Germany}
\vspace{-0.8cm}


\begin{abstract}

Spatiotemporal action localization requires the incorporation of two sources of information into the designed architecture: (1) temporal information from the previous frames and (2) spatial information from the key frame. Current state-of-the-art approaches usually extract these information with separate networks and use an extra mechanism for fusion to get detections. In this work, we present YOWO, a unified CNN architecture for real-time spatiotemporal action localization in video streams. YOWO is a single-stage architecture with two branches to extract temporal and spatial information concurrently and predict bounding boxes and action probabilities directly from video clips in one evaluation. Since the whole architecture is unified, it can be optimized end-to-end. The YOWO architecture is fast providing 34 frames-per-second on 16-frames input clips and 62 frames-per-second on 8-frames input clips, which is currently the fastest state-of-the-art architecture on spatiotemporal action localization task. Remarkably, YOWO outperforms the previous state-of-the art results on \mbox{J-HMDB-21} and \mbox{UCF101-24} with an impressive improvement of $\sim$3\% and $\sim$12\%, respectively. Moreover, YOWO is the first and only single-stage architecture that provides competitive results on AVA dataset. We make our code and pretrained models publicly available$^1$.
\end{abstract}


\fntext[urlnote]{https://github.com/wei-tim/YOWO}

\end{frontmatter}


\section{Introduction}

The topic of spatiotemporal human action localization has been spotlighted in recent years, which aims to not only recognize the occurrence of an action but also localize it in both time and space. In such a task, comparing with object detection in static images, temporal information plays an essential role. Finding an efficient strategy to aggregate spatial as well as temporal features makes the problem even more challenging. On the other hand, real-time human action detection is becoming increasingly crucial in numerous vision applications, such as human-computer interaction (HCI) systems, unmanned aerial vehicle (UAV) monitoring, autonomous driving, and urban security systems. Therefore, it is desirable and worthwhile to explore a more efficient framework to tackle this problem.



Inspired by the remarkable object detection architecture Faster R-CNN \cite{ren2015faster}, most state-of-the-art works \cite{hou2017tube, peng2016multi} extend the classic two-stage network architecture to action detection, where a number of proposals are produced in the first stage, then classification and localization refinement are performed in the second stage. However, these two-stage pipelines have three main shortcomings in the spatiotemporal action localization task. Firstly, the generation of action tubes which consist of bounding boxes across frames is much more complicated and time-consuming than 2D case. The classification performance is extremely dependent on these proposals, where the detected bounding boxes might be sub-optimal for the following classification task. Secondly, the action proposals focus only on the features of humans in the video, neglecting the relationship between humans and some attributes in the background, which yet is able to provide considerably crucial context information for action prediction. The third problem of a two-stage architecture is that training the region proposal network and the classification network separately does not guarantee to find the global optimum. Instead, only local optimum from the combination of two stages can be found. The training cost is also higher than single-stage networks, hence it takes longer time and needs more memory. 

\begin{figure}[t!]
	\centering
	\includegraphics[width = 1.0\linewidth]{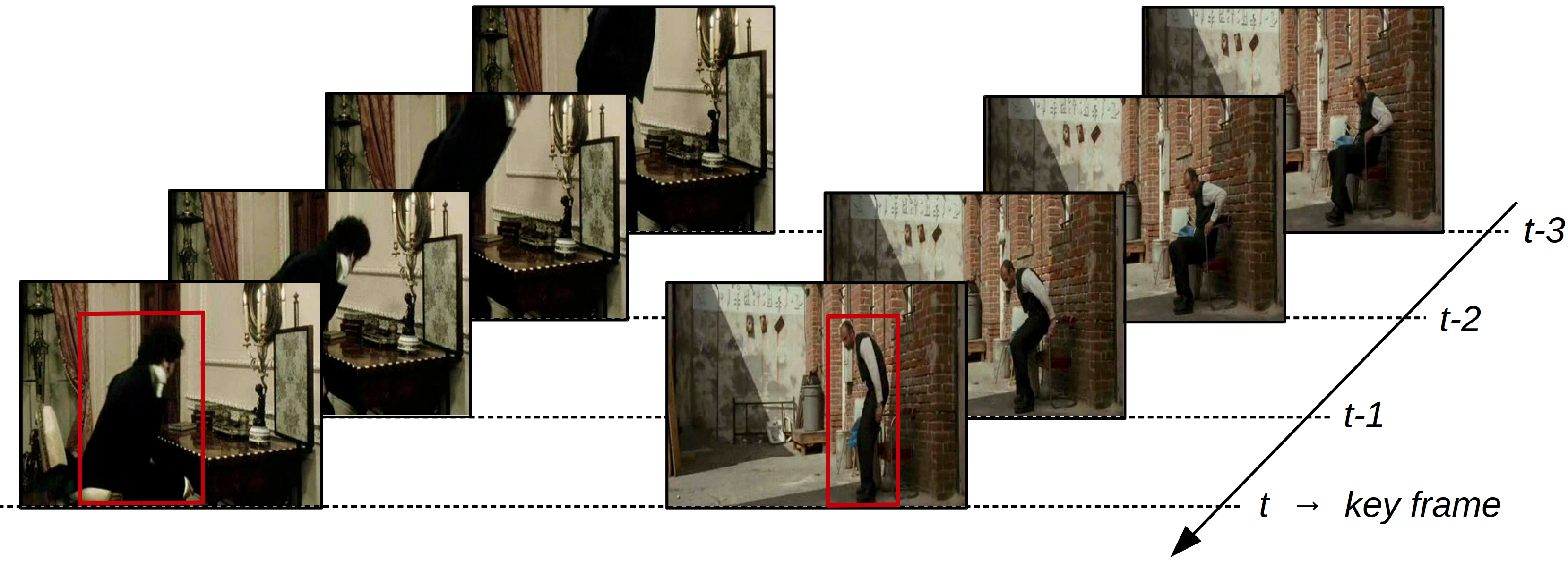}
	\caption{Standing or sitting? Although the person can be successfully detected, correct classification of the action cannot be made by looking only at the key frame. Temporal information from previous frames needs to be incorporated in order to understand if the person is sitting (left) or standing (right). Examples are from J-HMDB-21 dataset.}
	\label{fig:front_img}
\end{figure}

In this paper, we propose a novel single-stage framework, YOWO (You Only Watch Once), for spatiotemporal action localization in videos. YOWO prevents all of the three shortcomings mentioned above with a single-stage architecture. The intuitive idea of YOWO arises from human's visual cognitive system. For example, when we are absorbed into the story of a soap opera in front of the TV, each time our eyes capture a single frame. In order to understand which action each artist is performing, we have to relate current frame information (2D features from key frame) to the obtained knowledge from previous frames saved in our memory (3D features from clip). Afterwards, these two kinds of features are fused together to provide us with a reasonable conclusion. The example in Fig. \ref{fig:front_img} illustrates our inspiration.

YOWO architecture is a single-stage network with two branches. One branch extracts the spatial features of the key frame (i.e. current frame) via a 2D-CNN while the other branch models the spatiotemporal features of the clip consisting of previous frames via a 3D-CNN. To this end, YOWO is a causal architecture that can operate online on incoming video streams. In order to aggregate these 2D-CNN and 3D-CNN features smoothly, a channel fusion and attention mechanism is used, where we get the utmost out of inter-channel dependencies. Finally, we produce frame-level detections using the fused features, and provide a linking algorithm to generate action tubes.

In order to maintain real-time capability, we have operated YOWO on RGB modality. However, it must be noted that YOWO architecture is not restricted to operate only on RGB modality. Different branches can be inserted into YOWO for different modalities such as optical flow, depth etc. Moreover, in its 2D-CNN and 3D-CNN branches, any CNN architecture can be used according to the desired run-time performance, which is critical for real-world applications. 

YOWO operates with maximum 16 frames input since short clip lengths are necessary to achieve faster runtime for spatiotemporal action localization task. However, such small clip size is a limiting factor for the accumulation of temporal information. Therefore, we have made use of the long-term feature bank \cite{wu2019long} by extracting features with 3D-CNN for non-overlapping 8-frame clips for the whole videos using the trained 3D-CNN. Training of YOWO performed normally, but at inference time, we have averaged the 3D features centering the key-frame. This brought a considerable 6.9\% and 1.3\% frame-mAP increase on the final performance of the network. 

\vspace{0.3cm}
\textbf{Contributions} of this paper are summarized as follows:

\vspace{0.15cm}
\textbf{(i)} We propose a real-time single-stage framework for spatiotemporal action localization in video streams, named YOWO, which can be trained end-to-end with high efficiency. To the best of our knowledge, this is the first work which achieves bounding box regression on features extracted by a 2D-CNN and 3D-CNN, concurrently. These two kinds of features have a complementary effect to each other for the final bounding box regression and action classification. Moreover, we use a channel attention mechanism to aggregate the features smoothly from two branches above. We experimentally prove that channel-wise attention mechanism models the inter-channel relationship within the concatenated feature maps and boosts the performance significantly by fusing features more reasonably.

\vspace{0.15cm}
\textbf{(ii)} We perform a detailed ablation study on the YOWO architecture. We examined the effect of 3D-CNN, 2D-CNN, their aggregation and the fusion mechanism. Moreover, we have experimented different 3D-CNN architectures and different clip lengths to explore a further trade-off between the precision and speed.

\vspace{0.15cm}
\textbf{(iii)} YOWO is evaluated on AVA dataset, which is the first and only single-stage architecture that achieves competitive results compared to the state-of-the-art. Moreover, YOWO is the only causal architecture (i.e. future frames are not leveraged) that is evaluated on AVA dataset, hence can operate online.

\vspace{0.15cm}
\textbf{(iv)} We evaluate YOWO on \mbox{J-HMDB-21} and UCF101-24 benchmarks and establish new state-of-the-art results with an impressive 3.3\% and 12.2\% improvements on frame-mAP, respectively. Moreover, YOWO runs with 34 fps for 16-frames input clips and 62 fps for 8-frames input clips, which is the fastest state-of-the-art architecture available for spatiotemporal action localization task.

\section{Related Work}
{\bfseries Action recognition with deep learning.}
Since deep learning brings significant improvements in image recognition, numerous recent research efforts have been devoted to extend it for action recognition in videos. For action recognition, however, besides spatial features extracted from each individual image, temporal context across these frames also needs to be taken into account. Two-stream CNN is one effective strategy to extract spatial and temporal features separately and aggregate them together \cite{feichtenhofer2016convolutional} \cite{simonyan2014two} \cite{wang2016temporal}. Most of these works are based on optical flow, which requires significant computational power to extract, resulting in a time-consuming process. An alternative option to integrate CNN features over time is the implementation of recurrent networks, whose performance, however, is not so satisfying as recent CNN-based methods \cite{yue2015beyond}. \mbox{3D-CNNs} have been increasingly explored in video analysis tasks recently, which learns the features from both spatial and temporal dimensions simultaneously. 3D-CNNs are first exploited to extract spatiotemporal features in \cite{ji20123d}. Afterwards,  many 3D-CNN architectures have been proposed for action recognition task, such as C3D~\cite{tran2015learning}, I3D~\cite{carreira2017quo}, P3D~\cite{qiu2017learning}, R(2+1)D~\cite{tran2018closer}, SlowFast~\cite{feichtenhofer2019slowfast}, etc. In~\cite{hara2018can}, the effect of dataset size on performance is investigated for several 3D-CNN architectures. It must be noted that 3D-CNN architectures have much more parameters compared to 2D-CNNs, making them computationally expensive. In \cite{kopuklu2019resource}, 3D versions of some famous resource efficient CNN architectures are investigated. For resource efficiency, some other works focus on learning 2D features from single images with a 2D-CNN and then fusing them together to learn temporal features with a 3D-CNN \cite{zolfaghari2018eco}.

{\bfseries Spatiotemporal action localization.}
For object detection in images, \mbox{R-CNN} series extract region proposals using selective search \cite{girshick2014rich} or Region Proposal Network (RPN) \cite{ren2015faster} in the first stage and classify the objects in these potential regions in the second stage. Although Faster R-CNN \cite{ren2015faster} achieves state-of-the-art results in object detection, it is hard to implement it for real-time tasks due to its time-consuming two-stage architecture. Meanwhile, YOLO \cite{redmon2016you} and SSD \cite{liu2016ssd} aim to simplify this process to one stage and have outstanding real-time performance. For action localization in videos, due to the success of R-CNN series most of the research approaches propose first detecting the humans in each frame and then linking these bounding boxes reasonably as action tubes \cite{gkioxari2015finding, peng2016multi, hou2017tube}. Two-stream detectors introduce an additional stream on the base of the original classifier for optical flow modality \cite{peng2016multi} \cite{saha2016deep} \cite{singh2017online}. Some other works produce clip tube proposals with 3D-CNNs and achieve regression as well as classification on the corresponding 3D features \cite{hou2017tube} \cite{saha2016deep}, thus region proposal is necessary for them. In a recent work \cite{duarte2018videocapsulenet}, authors propose a 3D capsule network for video action detection which can jointly perform pixel-wise action segmentation along with action classification. However, it is too expensive in terms of computational complexity and number of parameters since it is a U-Net \cite{ronneberger2015u} based 3D-CNN architecture.

{\bfseries Attention modules.}
Attention is an effective mechanism to capture long-range dependencies and has been attempted to be used in CNNs to boost the performance in image classification \cite{wang2017residual} \cite{chen2017sca} \cite{woo2018cbam} and scene segmentation \cite{fu2019dual}. Attention mechanism is implemented spatial-wise and channel-wise in these works, in which spatial attention addresses the inter-spatial relationship among features while channel attention enhances the most meaningful channels and weakens the others. As a channel-wise attention block, Squeeze-and-Excitation module \cite{hu2018squeeze} is beneficial to increase CNN's performance with little computational cost. On the other hand, for video classification tasks, non-local block \cite{wang2018non} takes spatio-temporal information into account to learn the dependencies of features across frames, which can be viewed as a self-attention strategy.

Different from previous works, we have proposed a novel, unified framework called YOWO for the task of spatio-temporal action localization. We name it as YOWO as we make use of a clip only once and detect the corresponding actions in the key frame. However, to avoid the complex optical flow computation, we use 2D features of the key frame and 3D features of the clip together. Afterwards, these two kinds of features are fused together carefully with the application of attention mechanism such that rich contextual relationships are well taken into account. 

\section{Methodology}

In this section, we first present YOWO's architecture in detail, which extracts 2D features from the key frame as well as 3D features from the input clip concurrently and aggregates them together. Afterwards the implementation of channel fusion and attention mechanism is discussed, which provides the essential performance boost. Finally we describe the details of the training process for the YOWO architecture and the improved bounding box linking strategy for generation of action tubes in untrimmed videos.

\subsection{YOWO architecture}

\begin{figure*}[t!]
	\centering
	\includegraphics[width = 1.0\textwidth]{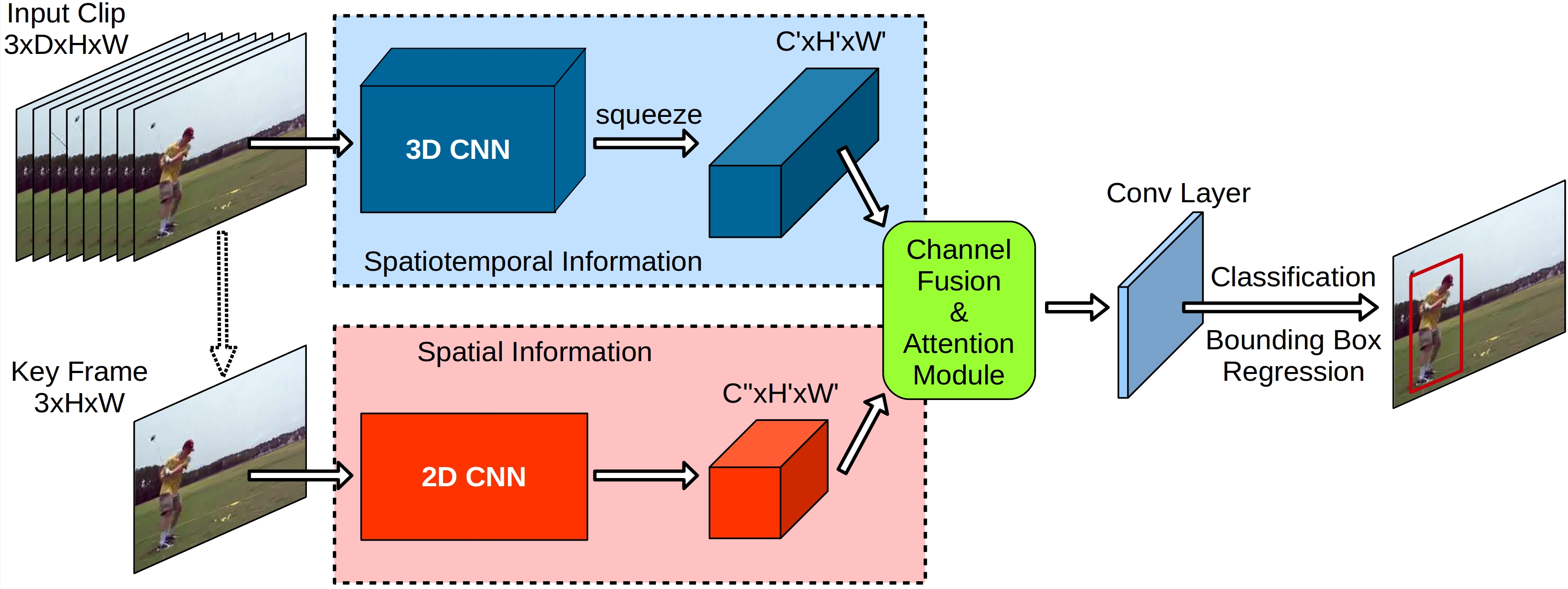}
	\caption{The YOWO architecture. An input clip and corresponding key frame is fed to a 3D-CNN and 2D-CNN to produce output feature volumes of $\left[C'' \times H' \times W'\right]$ and $\left[C' \times H' \times W'\right]$, respectively. These output volumes are fed to channel fusion and attention mechanism (CFAM) for a smooth feature aggregation. Finally, one last conv layer is used to adjust the channel number for final bounding box predictions.}
	\label{fig:yowo_arch}
	\vspace{0.5cm}
\end{figure*}

The YOWO architecture is illustrated in Fig.~\ref{fig:yowo_arch}, which can be divided into four major parts: 3D-CNN branch, 2D-CNN branch, CFAM and bounding box regression parts.

\vspace{0.2cm}
\noindent \textbf{3D-CNN Branch} Since contextual information is crucial for human action understanding, we utilize 3D-CNN to extract spatiotemporal features. 3D-CNNs are able to capture motion information by applying convolution operation not only in space dimension but also in time dimension. The basic 3D-CNN architecture in our framework is  \mbox{3D-ResNext-101} due to its high performance in Kinetics dataset \cite{hara2018can}. In addition to \mbox{3D-ResNext-101}, we have also experimented with different 3D-CNN models in our ablation study. For all 3D-CNN architectures, all of the layers after the last conv layer are discarded. The input to the 3D network is a clip of a video, which is composed of a sequence of successive frames in time order, and has a shape of $\left[C \times D \times H \times W\right]$, while the last conv layer of 3D ResNext-101 outputs a feature map of shape $\left[C' \times D' \times H' \times W'\right]$ where $C~=~3$, $D$ is the number of input frames, $H$ and $W$ are height and width of input images, $C'$ is the number of output channels, $D'~=~1$, $H'~=~\frac{H}{32}$ and $W'~=~\frac{W}{32}$. The depth dimension of the output feature map is reduced to 1 such that output volume is squeezed to $\left[C' \times H' \times W'\right]$  in order to match the output feature map of 2D-CNN.

\begin{figure*}[t!]
	\centering
	\includegraphics[width=1.0\textwidth]{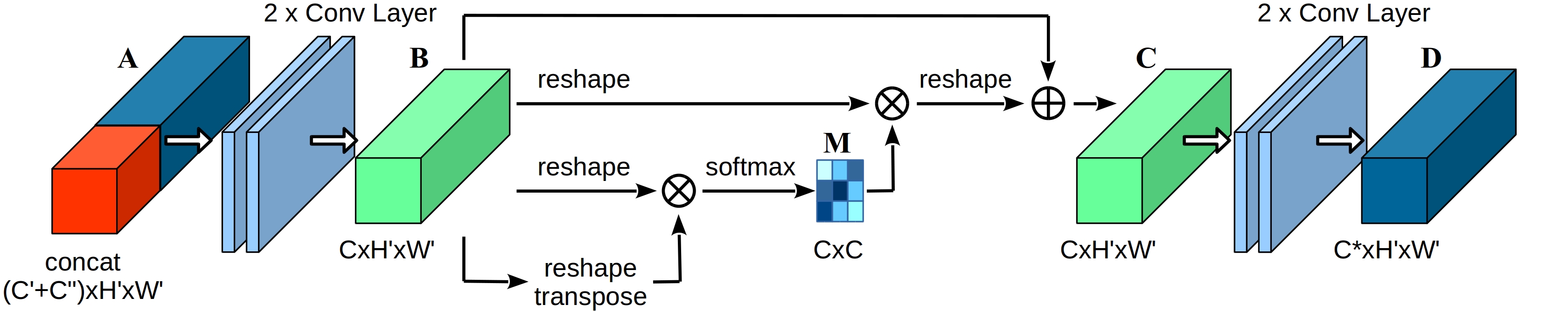}
	\caption{Channel fusion and attention mechanism for aggregating output feature maps coming from 2D-CNN and 3D-CNN branches.}
	\vspace{0.5cm}
	\label{fig:ch-attention}
\end{figure*}

\vspace{0.2cm}
\noindent \textbf{2D-CNN Branch} In the meantime, to address the spatial localization problem, 2D features of the key frame are also extracted in parallel. We employ Darknet-19 \cite{redmon2017yolo9000} as the basic architecture in our 2D CNN branch due to its good balance between accuracy and efficiency. The key frame with the shape $\left[C \times H \times W\right]$ is the most recent frame of the input clip, thus there is no need for an additional data loader. The output feature map of Darknet-19 has a shape of $\left[C'' \times H' \times W'\right]$ where $C~=~3$, $C''$ is the number of output channels, $H'~=~\frac{H}{32}$ and $W'~=~\frac{W}{32}$ similar to the 3D-CNN case. 

Another important characteristic of YOWO is that architectures in 2D CNN and 3D CNN branches can be replaced by arbitrary CNN architectures, which makes it more flexible. YOWO is designed to be simple and effort-saving to switch models. It must be noted that although YOWO has two branches, it is a unified architecture and can be trained end-to-end.

\vspace{0.2cm}
\noindent \textbf{Feature aggregation: \small Channel Fusion and Attention Mechanism (CFAM)} \normalsize We make the outputs of both 3D and 2D networks are of the same shape in the last two dimensions such that these two feature maps can be fused easily. We fuse the two feature maps using concatenation which simply stacks the features along channels. As a result, the fused feature map encodes both motion and appearance information which we pass as input to the CFAM module, which is based on Gram matrix to map inter-channel dependencies. Although Gram matrix based attention mechanism is originally used for style transfer \cite{gatys2015neural} and recently in segmentation task \cite{fu2019dual}, such an attention mechanism is beneficial for fusing features coming from different sources reasonably, which improves the overall performance significantly.

Fig.~\ref{fig:ch-attention} illustrates the used CFAM module. The concatenated feature map $\mathbf{A} \in \mathbbm{R}^{(C' + C'') \times H \times W}$ can be regarded as an abrupt combination of 2D and 3D information, which neglects interrelationship between them. Therefore, we first feed $A$ into two convolutional layers to generate a new feature map $\mathbf{B}~\in~\mathbbm{R}^{C \times H' \times W'}$. Afterwards, several operations are performed on the feature map~$\mathbf{B}$. 


Assume $\mathbf{F} \in \mathbbm{R}^{C \times N}$ is the reshaped tensor from feature map $\mathbf{B}$, where \mbox{$N = H \times W$}, which means that features in every single channel is vectorized to one dimension: 
\begin{equation}
    \mathbf{B} \in \mathbbm{R}^{C \times H \times W} \xrightarrow{vectorization} \mathbf{F} \in \mathbbm{R}^{C \times N}
\end{equation}
Then a matrix product between $\mathbf{F} \in \mathbbm{R}^{C \times N}$ and its transpose $\mathbf{F}^\mathrm{T} \in \mathbbm{R}^{N \times C}$ is performed to produce Gram matrix $\mathbf{G} \in \mathbbm{R}^{C \times C}$, which indicates the feature correlations across channels \cite{gatys2015neural}:
\begin{equation}
    \mathbf{G}~=~\mathbf{F} \cdot \mathbf{F}^\mathrm{T} \quad with \quad G_{ij}~=~\sum_{k=1}^N F_{ik} \cdot F_{jk}
\end{equation}
\noindent where each element $G_{ij}$ in the Gram matrix $\mathbf{G}$ represents the inner product between the vectorized feature maps $i$~and~$j$. After computing the Gram matrix, a softmax layer is applied to generate channel attention map $\mathbf{M} \in \mathbbm{R}^{C \times C}$:
\begin{equation}
    M_{ij}~=~\dfrac{exp(G_{ij})}{\begin{matrix} \sum_{j=1}^C exp(G_{ij}) \end{matrix}}
\end{equation}
\noindent where $M_{ij}$ is a score measuring the $j^{th}$ channel's impact on the $i^{th}$ channel. Therefore $M$ summaries the inter-channel dependency of features given a feature map. To perform the impact of attention map to original features, a further matrix multiplication between $\mathbf{M}$ and $\mathbf{F}$ is carried out and the result is reshaped back to 3-dimensional space $\mathbbm{R}^{C \times H \times W}$, which has the same shape as the input tensor:
\begin{equation}
    \mathbf{F'}~=~\mathbf{M} \cdot \mathbf{F}
\end{equation}
\begin{equation}
    \mathbf{F'} \in \mathbbm{R}^{C \times N} \xrightarrow{reshape} \mathbf{F''} \in \mathbbm{R}^{C \times H \times W}
\end{equation}

The output of channel attention module $\mathbf{C} \in \mathbbm{R}^{C \times H \times W}$ combines this result with the original input feature map $\mathbf{B}$ with a trainable scalar parameter $\alpha$ using an element-wise sum operation, and $\alpha$ gradually learns a weight from $0$:
\begin{equation} \label{eq:attention}
    \mathbf{C}~=~\alpha \cdot \mathbf{F''} + \mathbf{B}
\end{equation}
The Eq. (\ref{eq:attention}) shows that the final feature of each channel is a weighted sum of the features of all channels and original features, which models the long-range semantic dependencies between feature maps. Finally, the feature map \mbox{$\mathbf{C} \in  \mathbbm{R}^{C \times H' \times W'}$} is fed into two more convolutional layers to generate the output feature map \mbox{$\mathbf{D} \in \mathbbm{R}^{C^* \times H' \times W'}$} of the CFAM module. Two convolutional layers at the beginning and the end of CFAM modules contain utmost importance since they help to mix the features coming from different backbones and having possibly different distributions. Without these convolutional layers, CFAM marginally improves the performance. 

Such an architecture promotes the feature representativeness in terms of inter-dependencies among channels and thus the features from different branches can be aggregated reasonably and smoothly. Besides, Gram matrix takes the whole feature map into consideration, where the dot product of each two flattened feature vectors presents the information about the relation between them. A larger product indicates that the features in these two channels are more correlated while a smaller product suggests that they are different from each other. For a given channel, we allocate more weights to the other channels  which are much correlated and have more impact to it. By means of this mechanism, contextual relationship is emphasized and feature discriminability is enhanced.

\vspace{0.2cm}
\noindent \textbf{Bounding box regression} We follow the same guidelines of YOLO \cite{redmon2017yolo9000} for bounding box regression. A final convolutional layer with $1 \times 1$ kernels is applied to generate desired number of output channels. For each grid cell in $H' \times W'$, 5 prior anchors are selected by k-means technique on corresponding datasets with \textit{NumCls} class conditional action scores, 4 coordinates and confidence score making the final output size of YOWO $\left[(5 \times (NumCls + 5)) \times H' \times W'\right]$. The regression of bounding boxes are then refined based on these anchors.

We have used input resolution of \mbox{$224 \times 224$} for both training and testing time. Applying multi-scale training with different resolutions has not shown any performance improvement in our experiments.  The loss function is defined similar to the original YOLOv2 network \cite{redmon2017yolo9000} except that we apply smooth L$_1$ loss with beta=1 for localization as in \cite{girshick2015fast}, which is given as follows:
\begin{equation}
    L_{1,smooth}(x,y) =  
    \begin{cases} 
    0.5(x-y)^2  &   if |x-y|<1 \\
    |x-y|-0.5   &   otherwise
    \end{cases}
\end{equation}
where x and y refers to network prediction and ground truth, respectively. Smooth L$_1$ loss is less sensitive to outliers than the MSE loss and prevents exploding gradients in some cases. We still apply the MSE loss for confidence score, which is given as follows:
\begin{equation}
    L_{MSE}(x,y) = (x-y)^2
\end{equation}
The final detection loss becomes the summation of individual coordinate losses for x, y, width and height; and confidence score loss, which is given as follows:
\begin{equation}
    L_D = L_x + L_y + L_w + L_h + L_{conf}
\end{equation}
We have applied focal loss \cite{lin2017focal} for classification, which is given as follows:
\begin{equation}
    L_{focal}(x,y) = y(1-x)^\gamma log(x) + (1-y)x^\gamma log(1-x)
\end{equation}
where x is the softmaxed network prediction and y $\in \{0,1\}$ is the ground truth class label. $\gamma$ is the modulating factor, which reduce the loss of samples with high confidence (i.e. easy samples) and increase the loss of samples with low confidence (i.e. hard samples). However, AVA dataset is a multi-label dataset where each person performs one pose action (e.g. walking, standing, etc.) and multiple human-human or human-object interaction actions. Therefore, we have applied softmax to pose classes and sigmoid to the interaction actions. Moreover, AVA is an unbalanced dataset and modulating factor $\gamma$ is not enough to tackle dataset imbalance. Therefore, we have used $\alpha$-balanced variant of focal loss \cite{lin2017focal}. For $\alpha$ term, we have used exponential of class sample ratios.

The final loss that is used for the optimization of YOWO architecture is the summation of detection and classification loss, which is given as follows:
\begin{equation}
    L_{final} = \lambda L_D + L_{Cls}
\end{equation}
where $\lambda = 0.5$ performs best in our experiments.

\subsection{Implementation details}
We initialize the 3D and 2D network parameters separately: 3D part with pretrained models on Kinetics \cite{carreira2017quo} and 2D part with pretrained models on PASCAL VOC \cite{lin2014microsoft}. Although our architecture consists of 2D-CNN and 3D-CNN branches, the parameters are able to be updated jointly. We select the mini-batch stochastic gradient decent algorithm with momentum and weight decay strategy to optimize the loss function. The learning rate is initialized as 0.0001 and reduced with a factor of 0.5 after 30k, 40k, 50k and 60k iterations. For the dataset UCF101-24, the training process is completed after 5 epochs while for J-HMDB-21 after 10 epochs. The complete architecture is implemented and trained end-to-end in PyTorch using a single Nvidia Titan XP GPU.

In the trainings, because of the small number of samples in J-HMDB-21, we freeze all the 3D conv network parameters thus the convergence is faster and over-fitting risk can be reduced. In addition, we deploy several data augmentation techniques at training time such as flipping images horizontally in the clip, random scaling and random spatial cropping. During testing, only detected bounding boxes with confidence score larger than threshold 0.25 are selected and then post-processed with non-maximum suppression with a \mbox{threshold of 0.4} for UCF101-24 and J-HMDB-21 datasets; and 0.5 for AVA dataset.

\subsection{Linking Strategy}
As we have already obtained frame-level action detections, next step is to link these detected bounding boxes to construct action tubes in the whole video for UCF101-24 and J-HMDB-21 datasets. We make use of the linking algorithm described in \cite{gkioxari2015finding, peng2016multi} to find the optimal video-level action detections.

Assume $R_t$ and $R_{t+1}$ are two regions from consecutive frames \textit{t} and \textit{t+1}, the linking score for an action class $c$ is defined as
\begin{equation}
\begin{aligned}
s_c(R_t, R_{t+1})~=~\ & \psi(ov) \cdot [ s_c(R_t) + s_c(R_{t+1}) \\
& + \alpha \cdot s_c(R_t) \cdot s_c(R_{t+1}) \\
& + \beta \cdot ov(R_t, R_{t+1}) ]
\end{aligned}
\end{equation}
where $s_c(R_t)$, $s_c(R_{t+1})$ are class specific scores of regions $R_t$ and $R_{t+1}$, $ov$ is the intersection-over-union of these two regions, $\alpha$ and $\beta$ are scalars. $\psi(ov)$ is a constraint which is equal to 1 if an overlap exists ($ov > 0$), otherwise $\psi(ov)$ is equal to 0. We extend the linking score definition in \cite{peng2016multi} with an extra element $\alpha \cdot s_c(R_t) \cdot s_c(R_{t+1})$, which takes the dramatic change of scores between two successive frames into account and is able to improve the performance of video detection in experiments. After all the linking scores are computed, Viterbi algorithm is deployed to find the optimal path to generate action tubes.

\subsection{Long-Term Feature Bank}

Although YOWO's inference is online and causal with small clip size, \mbox{16-frame} input limits the temporal information required for action understanding. Therefore, we make use of a long-term feature bank (LFB) similar to \cite{wu2019long}, which contains features coming from 3D backbone at different timestamps. At inference time, 3D features centering the key-frame are averaged and the resulting feature map is used as input to the CFAM block. LFB features are extracted for non-overlapping 8-frame clips using the pretrained 3D ResNeXt-101 backbone. We have used 8 features (if available) centering the key-frame. So, total number of 64 frames are utilized at inference time. Utilization of LFB increases action classification performance similar to difference between clip accuracy and video accuracy in video datasets. However, introduction of LFB makes the resulting architecture non-causal since future 3D features are used at inference time. 
\section{Experiments}
To evaluate YOWO's performance, three popular and challenging action detection datasets are selected: UCF101-24 \cite{soomro2012ucf101}, J-HMDB-21 \cite{jhuang2013towards} and AVA \cite{gu2018ava}. Each of these datasets contains different characteristics, which are compared in Table~\ref{tab:datasets_comparison}. We follow the official evaluation metrics strictly to report the results and compare the performance of our method with the state-of-the-art. 

\begin{table}[t!]
    \begin{adjustbox}{width=\textwidth}
    \begin{tabular}{|c|c|c|c|c|}
    \hline
    \multicolumn{1}{|c|}{\textbf{Dataset}} & \multicolumn{1}{c|}{\begin{tabular}[c]{@{}c@{}}\textbf{\# of labelled}\\ \textbf{person per frame}\end{tabular}} & \multicolumn{1}{c|}{\begin{tabular}[c]{@{}c@{}}\textbf{\# of labels}\\ \textbf{per person}\end{tabular}} & \multicolumn{1}{c|}{\begin{tabular}[c]{@{}c@{}}\textbf{background}\\ \textbf{people}\end{tabular}}   & \multicolumn{1}{c|}{\begin{tabular}[c]{@{}c@{}}\textbf{densely}\\ \textbf{annotated}\end{tabular}} \\ \hline
    \textbf{UCF101-24}   &  one or two  &  one         & \cmark    & \xmark   \\ \hline
    \textbf{J-HMDB-21}   &  one         &  one         & \xmark    & \cmark   \\ \hline
    \textbf{AVA}         &  multiple    &  multiple    & \xmark    & \cmark*   \\ \hline
    \end{tabular}
    \end{adjustbox}
    \caption{Comparison of evaluated datasets. Background people refers that in there are people in some of the frames who are not annotated. *AVA is densely annotated with 1Hz rate.}
    \label{tab:datasets_comparison}
\end{table}

\vspace{0.2cm}
\subsection{Datasets and evaluation metrics}
\noindent{\bfseries UCF101-24} is a subset of UCF101 \cite{soomro2012ucf101}, which is originally an action recognition dataset of realistic action videos. UCF101-24 contains 24 action classes and 3207 videos, for which the corresponding spatiotemporal annotations are provided. In addition, there might be multiple action instances in each video, which have the same class label but different spatial and temporal boundaries. Such a property makes video-level action detection much more challenging. As in previous works, we perform all the experiments on the first split.

\vspace{0.2cm}
\noindent{\bfseries J-HMDB-21} is a subset of the HMDB-51 dataset \cite{kuehne2011hmdb} and consists of 928 short videos with 21 action categories in daily life. Each video is well trimmed and has a single action instance across all the frames. We report our experimental results on the first split.

\vspace{0.2cm}
\noindent{\bfseries AVA} is a video dataset of spatiotemporally localized Atomic Visual Actions (AVA). The AVA
dataset contains 80 atomic visual actions, which are densely annotated for 430 15-minute video clips, where actions are localized in space and time. This results in 1.58M action labels and each person is labelled with multiple actions. Following the guidelines of the ActivityNet challenge, we evaluate YOWO on most-frequent 60 action classes of AVA dataset. If not stated otherwise, we report results on version 2.2 of AVA dataset.

\vspace{0.2cm}
\noindent{\bfseries Evaluation metrics:} We employ two popular metrics used by the most researches in the region of spatio-temporal action detection to generate convincing evaluations. Following strictly the rule applied by the PASCAL VOC 2012 metric \cite{everingham2010pascal}, frame-mAP measures the area under the precision-recall curve of the detections for each frame. On the other hand, video-mAP focuses on the action tubes \cite{gkioxari2015finding}. If the mean per frame intersection-over-union with the ground truth across the frames of the whole video is greater than a threshold and in the meanwhile the action label is correctly predicted, then this detected tube is regarded as a correct instance. Finally, the average precision for each class is computed and the average over all classes is reported. For AVA dataset, we only use frame-mAP with Intersection of Union (IoU) threshold of 0.5 since annotations are sparsely provided with 1 Hz.

\subsection{Ablation study}

\vspace{0.2cm}
{\bfseries 3D network, 2D network or both?}
Depending only on its own, neither 3D-CNN nor 2D-CNN can solve the spatiotemporal localization task independently. However, if they operate simultaneously, there is potential to benefit from one another. Results on comparing the performance of different architectures are reported in Table~\ref{tab:different sctructures}. We first observe that a single 2D network can not provide a satisfying result since it does not take temporal information into account. A single 3D network is better at capturing motion information and the fusion of 2D and 3D networks (simple concatenation) can improve the performance by around 3\%, 6\% and 2\% mAP compared to 3D network on UCF101-24, J-HMDB-21 and AVA datasets, respectively. This indicates that 2D-CNN learns finer spatial features and 3D-CNN concentrates more on the motion process yet the spatial drift of an action in the clip may lead to a lower localization accuracy. It is also shown that CFAM module further boosts the performance from 73.8\% to 79.2\% on UCF101-24, from 47.1\% to 64.9\% on J-HMDB-21 and from 16.0\% to 16.4\% on AVA dataset. This clearly shows the importance of the attention mechanism which strengthens the inter-dependencies among channels and helps aggregating features more reasonably.

\begin{table}[t!]
    \centering
    \begin{tabular}{cccc}
        \specialrule{.1em}{.2em}{.2em}
        \textbf{Model} & \textbf{UCF101-24} & \textbf{J-HMDB-21} &  \textbf{AVA} \\
        \specialrule{.1em}{.2em}{.2em}
        2D                      & 61.6              & 36.0      & 13.2  \\
        3D                      & 70.5              & 41.5      & 13.7  \\ 
        2D + 3D                 & 73.8              & 47.1      & 16.0  \\ 
        2D + 3D + CFAM          & {\bfseries 79.2}  & \textbf{64.9}     & \textbf{16.4}  \\
        \specialrule{.1em}{.2em}{.2em}
    \end{tabular}
    \caption{Frame-mAP @ IoU 0.5 results on UCF101-24, J-HMDB-21 and AVA datasets for different models. For all architectures, the input to 3D-CNNs is 8 frames clips~with~downsampling~of~1.}
	\label{tab:different sctructures}
\end{table}

Moreover, in order to explore the impact of each 2D-CNN, 3D-CNN and CFAM blocks, we investigate the localization and the classification performance of different architectures, which is given in Table~\ref{tab:loc_clsf}. For localization, we look at the recall value, which is the ratio of the number of correctly localized actions to the total number of ground truth actions. For classification, we look at the classification accuracy of the correctly localized detections. For this analysis, we have excluded AVA dataset since it contains multiple actions per person, hence classification accuracy cannot be calculated. For both UCF101-24 and J-HMDB-21 datasets, 2D network is better at localization while 3D network performs better at classification. It is also obvious that CFAM module boosts both localization and classification performance.

\begin{table}[t!]
    \centering
    \begin{tabular}{cccc}
        \specialrule{.1em}{.2em}{.2em}
        & \textbf{Model} & \phantom{aa}\begin{tabular}[c]{@{}l@{}}\textbf{Localization} \\ \hspace{0.4cm}\textbf{(recall)}\end{tabular}\phantom{aa} & \phantom{aa}\textbf{Classif.}\phantom{aa} \\ 
        \specialrule{.1em}{.2em}{.6em}
        \multicolumn{1}{c|}{\multirow{4}{*}{\rotatebox[origin=c]{90}{\textbf{UCF101-24}}}}  
        & \phantom{a}2D   & 91.7   & 85.9 \\ 
        \multicolumn{1}{c|}{}             & \phantom{a}3D              & 90.8   & 92.9   \\ 
        \multicolumn{1}{c|}{}             & \phantom{a}2D + 3D         & 93.2    & 93.7    \\ 
        \multicolumn{1}{c|}{}             & \phantom{a}2D + 3D + CFAM  & \textbf{93.5}   & \textbf{94.5}   \\
        \specialrule{.1em}{.4em}{.4em}
        \multicolumn{1}{c|}{\multirow{4}{*}{\rotatebox[origin=c]{90}{\textbf{J-HMDB-21}}}}    & \phantom{a}2D   & 94.3  & 50.6  \\ 
        \multicolumn{1}{c|}{}     & \phantom{a}3D                      & 76.3    & 69.3   \\ 
        \multicolumn{1}{c|}{}     & \phantom{a}2D + 3D                 & 94.5    & 63.0   \\ 
        \multicolumn{1}{c|}{}     & \phantom{a}2D + 3D + CFAM          & \textbf{97.3}    & \textbf{76.1}   \\
        \specialrule{.1em}{.6em}{.2em}
    \end{tabular}
    \caption{Localization @ IoU 0.5 (recall) and classification results on UCF101-24 and J-HMDB-21 datasets. For all architectures, the input to 3D-CNNs is 8 frames clips with downsampling of 1.}
	\label{tab:loc_clsf}
\end{table}

\begin{figure*}[b!]
    \centering
    \includegraphics[width=0.90\linewidth,keepaspectratio]{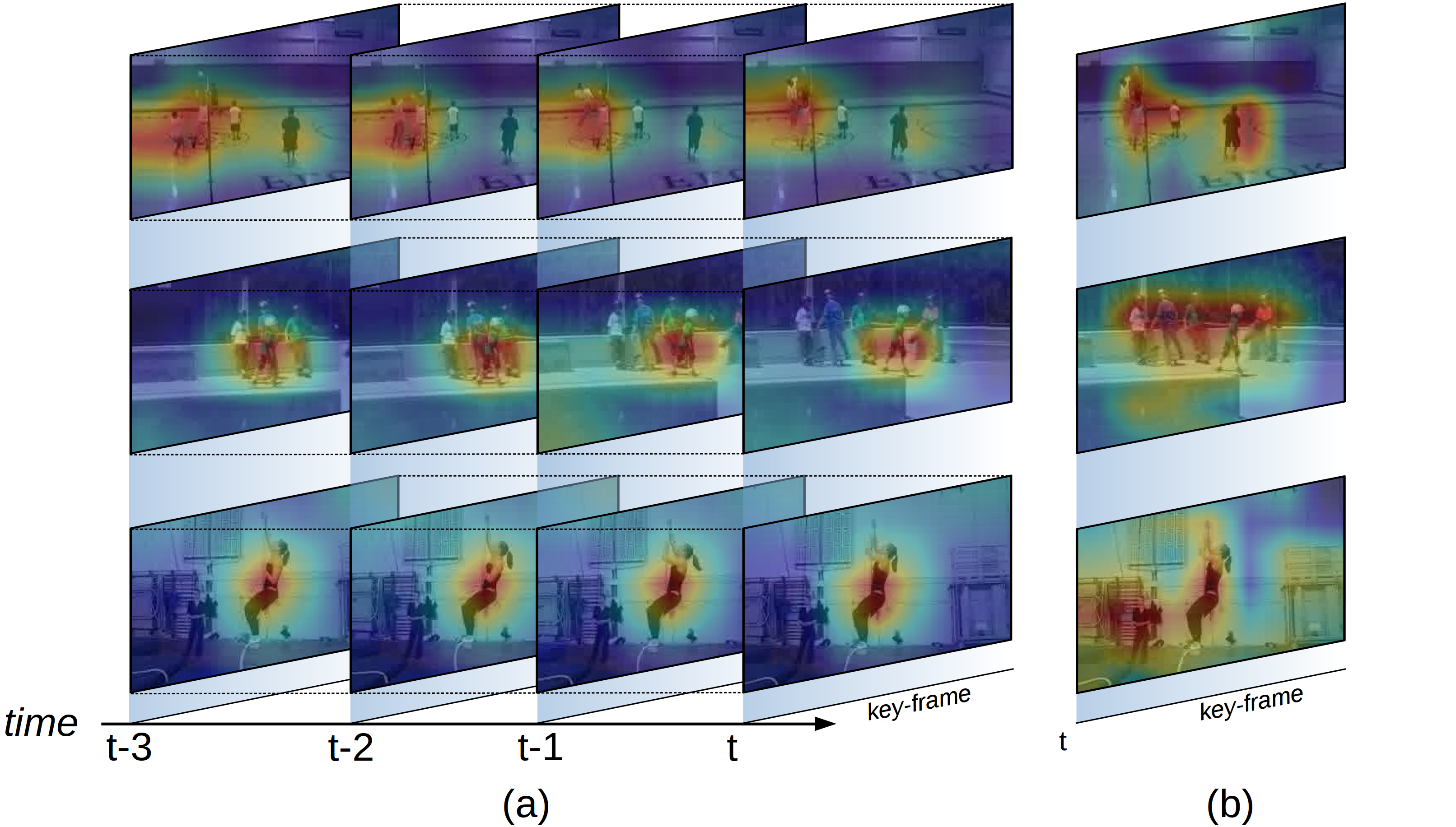}
    \caption{Activation maps for (a) 3D-CNN backbone and (b) 2D-CNN backbone. 3D-CNN backbone focuses on areas where there is a movement/action happening, whereas 2D-CNN backbone focuses on all the people in the key-frame. Examples are volleyball spiking (top), skate boarding (middle) and rope climbing (bottom).}
    \label{fig:attention_maps}
\end{figure*}

We have also visualized the activations maps \cite{zhou2015cnnlocalization} for 2D and 3D backbones of the trained model, which is shown in Fig. \ref{fig:attention_maps}. Conforming our findings in Table \ref{tab:loc_clsf}, 3D backbone focuses on the parts of the clip where a motion is occurring and 2D backbone focuses on fine spatial information on complete body parts of people. This validates that backbones of YOWO extract complementary features.   

\begin{table}[t!]
    \centering
    \begin{tabular}{cccc}
        \specialrule{.1em}{.2em}{.2em}
        \textbf{Input} & \textbf{UCF101-24} & \textbf{J-HMDB-21 } & \textbf{AVA} \\
        \specialrule{.1em}{.2em}{.2em}
        8-frames (d=1)      & 79.2  & 64.9  & 16.4 \\
        8-frames (d=2)      & 78.5  & 61.5  & 16.1 \\
        8-frames (d=3)      & 78.4  & 61.0  & 16.0 \\
        \specialrule{.04em}{.10em}{.15em}
        16-frames (d=1)     & 80.4  & \textbf{74.4}    & 17.9\\ 
        16-frames (d=2)     & 79.0  & 71.4  & 17.2 \\
        \specialrule{.1em}{.2em}{.2em}
    \end{tabular}
    \caption{Frame-mAP @ IoU 0.5 results on UCF101-24, J-HMDB-21 and AVA datasets for different clip lengths and different downsampling rates \textit{d}.}
	\label{tab:temporal}
\end{table}

\vspace{0.2cm}
{\bfseries How many frames are suitable for temporal information?}
For 3D-CNN branch, different clip lengths with different downsampling rates can change the performance of overall YOWO architecture \cite{kopuklu2018analysis}. Therefore, we conduct experiments with 8-frames and 16-frames clips with different downsampling rates, which is given in Table~\ref{tab:temporal}. For example, 8-frames (d=3) refers to selecting 8 frames from 24 frames window with downsampling rate of 3. Specifically, we compare three downsampling rates $d = 1, 2, 3$ for clip length 8-frames and two downsampling rates $d = 1, 2$ for 16-frames clip length. As expected, we observe that the framework with input of $16$ frames performs better than $8$ frames since long frame sequence contains more temporal information. However, as downsampling rate is increased, the performance becomes worse. We conjecture that downsampling hinders capturing motion patterns properly and too long sequence may break the temporal contextual relationship. Especially for some quick motion classes, a long sequence may contain several unrelated frames, which can be viewed as noise.

\begin{table}[t!]
    \centering
    \begin{tabular}{lcccc}
        \specialrule{.1em}{.2em}{.2em}
        \multicolumn{1}{c}{\multirow{2}{*}{\textbf{Model}}} & \multicolumn{1}{c}{\multirow{2}{*}{\textbf{GFLOPs}}} & \multicolumn{3}{c}{\textbf{Frame-mAP (@ IoU 0.5)}}  \\ \cline{3-5} \addlinespace
        \multicolumn{1}{c}{}                       & \multicolumn{1}{c}{}                       & 
        \multicolumn{1}{c}{\textbf{UCF101-24}} &
        \multicolumn{1}{c}{\textbf{J-HMDB-21}} &
        \multicolumn{1}{c}{\textbf{AVA}} \\ 
        \specialrule{.1em}{.2em}{.2em}
        3D-ResNext-101           & 38.6      & \textbf{80.4}     & \textbf{74.4}    & \textbf{17.9} \\
        \specialrule{.04em}{.10em}{.15em}
        3D-ResNet-101            & 54.7      & 78.1     & 70.8  & 17.7     \\
        3D-ResNet-50             & 39.3      & 77.8     & 61.3  & 17.1     \\
        3D-ResNet-18             & 33.3      & 72.6     & 57.5  & 15.1     \\
        \specialrule{.04em}{.10em}{.15em}
        3D-ShuffleNetV1 2.0x     & 2.2       & 71.3     & 54.8  & 16.1  \\
        3D-ShuffleNetV2 2.0x     & 1.8       & 71.4     & 55.3  & 15.1  \\ 
        3D-MobileNetV1 2.0x      & 2.6       & 67.3     & 48.5  & 14.9  \\
        3D-MobileNetV2 1.0x      & 2.2       & 66.6     & 52.5  & 16.1  \\
        \specialrule{.1em}{.2em}{.2em}
    \end{tabular}
    \caption{Performance comparison of different 3D backbones on UCF101-24, \mbox{J-HMDB-21} and AVA datasets. For all architectures, Darknet-19 is used as 2D backbone. The number of floating point operation (FLOPs) are calculated for corresponding 3D backbones for 16 frames (d=1) clips with spatial resolution of 224 $\times$ 224.}
    \vspace{0.2cm}
	\label{tab:dif_3dcnn}
\end{table}

\vspace{0.2cm}
{\bfseries Is it possible to save model complexity with more efficient networks?}
We have chosen 3D-ResNext-101 \cite{hara2018can} since it has multiple cardinalities thus is able to learn more complicated features. However, it is a heavy backbone with a huge number of parameters and computational complexity. Therefore, we have replaced the 3D backbone with 3D-ResNet for different depths and with some other resource efficient 3D-CNN architectures \cite{kopuklu2019resource}. Table~\ref{tab:dif_3dcnn} reports the achieved performance on all three datasets together with the number of floating point operations (FLOPs) for each 3D backbone. We find that even with light-weight architecture in 3D backbones, our framework is still better than 2D network. However, Table~\ref{tab:dif_3dcnn} clearly shows the importance of the 3D backbone. The stronger 3D-CNN architecture we use, better the achieved results.

\subsection{State-of-the-art comparison}

We have compared YOWO with other state-of-the-art architectures on J-HMDB-21, UCF101-24 and AVA datasets. For the sake of fairness, we have excluded VideoCapsuleNet \cite{duarte2018videocapsulenet} as it uses different video-mAP calculation without constructing action tubes via some linking strategies. However, YOWO still performs around 9\% and 8\% better than VideoCapsuleNet in terms of frame-mAP @ 0.5 IoU on J-HMDB-21 and UCF101-24, respectively.

\begin{table}[t!]
    \centering
    \begin{tabular}{lcccc}
        \specialrule{.1em}{.2em}{.2em}
        \multicolumn{1}{c}{\multirow{2}{*}{\textbf{Method}}} & \multicolumn{1}{c}{\multirow{2}{*}{\phantom{aa}\textbf{Frame-mAP}\phantom{aa}}} & \multicolumn{3}{c}{\textbf{Video-mAP}}                                   \\ \cline{3-5} \addlinespace
        \multicolumn{1}{c}{}                       & \multicolumn{1}{c}{}                       & 
        \multicolumn{1}{c}{\phantom{aa}\textbf{0.2}\phantom{aa}} &
        \multicolumn{1}{c}{\phantom{aa}\textbf{0.5}\phantom{aa}} &
        \multicolumn{1}{c}{\phantom{aa}\textbf{0.75}\phantom{aa}} \\
        \specialrule{.1em}{.2em}{.2em}
        Peng w/o MR \cite{peng2016multi}   & 56.9   & 71.1   & 70.6   & 48.2   \\ 
        Peng w/ MR \cite{peng2016multi}   & 58.5   & 74.3   & 73.1   & -   \\ 
        ROAD \cite{singh2017online}   & -   & 73.8   & 72.0   & 44.5   \\
        T-CNN \cite{hou2017tube}   & 61.3   & 78.4   & 76.9   & -   \\
        ACT \cite{kalogeiton2017action}   & 65.7   & 74.2   & 73.7   & 52.1   \\
        P3D-CTN \cite{wei2019p3d}   & 71.1   & 84.0   & 80.5   & -   \\
        TPnet \cite{singh2018predicting}  & -   & 74.8   & 74.1   & {\bfseries 61.3}   \\
        \specialrule{.04em}{.10em}{.15em}
        \textbf{YOWO (16-frame)}   & 74.4   & 87.8   & 85.7   & 58.1   \\
        \specialrule{.04em}{.10em}{.15em}
        \textbf{YOWO+LFB*}   & {\bfseries 75.7}   & \textbf{88.3}   & {\bfseries 85.9}   & 58.6   \\
        \specialrule{.1em}{.2em}{.2em}
    \end{tabular}
    \caption{Comparison with state-of-the-art methods on the J-HMDB-21 dataset. Results are reported for frame-mAP under IoU threshold of 0.5 and video-mAP under different IoU thresholds. * version of YOWO is non-causal.}
	\label{tab:jhmdb21_sota}
\end{table}

\vspace{0.2cm}
\noindent \textbf{Performance comparison on J-HMDB-21 dataset.} YOWO is compared with the previous state-of-the-art methods on J-HMDB-21 in Table~\ref{tab:jhmdb21_sota}. Using the standard metrics, we report the frame-mAP at IoU threshold of $0.5$ and the video-mAP at various IoU thresholds. YOWO (16-frame) consistently outperforms the state-of-the-art results on dataset J-HMDB-21, with a frame-mAP improvement of $3.3\%$ and a video-mAP improvement of $3.8\%$, $5.2\%$ at IoU thresholds of $0.2$ and $0.5$, respectively. Utilization of LFB brings further improvements on the performance. However, this improvement is marginal since the video duration of videos of J-HMDB-21 dataset is maximum 40 frames.  

\vspace{0.2cm}
\noindent \textbf{Performance comparison on UCF101-24 dataset.} Table~\ref{tab:ucf101-24_sota} presents the comparison of YOWO with the state-of-the-art methods on UCF101-24. YOWO (16-frame) achieves $80.4\%$ with respect to frame-mAP metric, which is significantly better than the others by preceding the second best result with $5.4\%$ improvement. As for video-mAP, our framework also produces very competitive results even though we just utilize a simple linking strategy. Utilization of LFB brings considerable improvement this time since the duration of UCF101-24 videos is much bigger than J-HMDB-21 videos. LFB further increases frame-mAP performance by around $7\%$.

\begin{table}[t!]
    \centering
    \begin{tabular}{lcccc}
        \specialrule{.1em}{.2em}{.2em}
        \multicolumn{1}{c}{\multirow{2}{*}{\textbf{Method}}} & \multicolumn{1}{c}{\multirow{2}{*}{\phantom{aa}\textbf{Frame-mAP}\phantom{aa}}} & \multicolumn{3}{c}{\textbf{Video-mAP}}                                   \\ \cline{3-5} \addlinespace
        \multicolumn{1}{c}{}                       & \multicolumn{1}{c}{}                       & 
        \multicolumn{1}{c}{\phantom{aa}\textbf{0.1}\phantom{aa}} &
        \multicolumn{1}{c}{\phantom{aa}\textbf{0.2}\phantom{aa}} &
        \multicolumn{1}{c}{\phantom{aa}\textbf{0.5}\phantom{aa}} \\
        \specialrule{.1em}{.2em}{.2em}
        Peng w/o MR \cite{peng2016multi}   & 64.8   & 49.5   & 41.2   & -   \\ 
        Peng w/ MR \cite{peng2016multi}   & 65.7   & 50.4   & 42.3   & -   \\ 
        ROAD \cite{singh2017online}   & -   & -   & 73.5   & 46.3    \\
        T-CNN \cite{hou2017tube}   & 41.4   & 51.3   & 47.1   & -   \\
        ACT \cite{kalogeiton2017action}   & 69.5   & -   & 77.2   & 51.4   \\
        MPS \cite{alwando2018cnn}   & -   & 82.4   & 72.9   & 41.1   \\
        STEP \cite{yang2019step}   & 75.0   & 83.1   & 76.6   & -   \\
        \specialrule{.04em}{.10em}{.15em}
        \textbf{YOWO (16-frame)}   & 80.4   & 82.5   & 75.8   & 48.8   \\
        \specialrule{.04em}{.10em}{.15em}
        \textbf{YOWO+LFB*}   & {\bfseries 87.3}   & \textbf{86.1}   & {\bfseries 78.6}   & \textbf{53.1}   \\
        \specialrule{.1em}{.2em}{.2em}
    \end{tabular}
    \caption{Comparison with state-of-the-art methods on the UCF101-24 dataset. Results are reported for frame-mAP under IoU threshold of 0.5 and video-mAP under different IoU thresholds. * version of YOWO is non-causal.}
	\label{tab:ucf101-24_sota}
\end{table}

\begin{table}[t!]
    \centering
    \begin{adjustbox}{width=\textwidth}
    \begin{tabular}{l|c|c|c|c|c}
        \specialrule{.1em}{.06em}{.06em}
        \textbf{Method} & \begin{tabular}[c]{@{}l@{}}\textbf{Single} \\ \hspace{0.1cm}\textbf{Stage}\end{tabular} & \textbf{Input} & \textbf{AVA} & \textbf{Pretrain} & \textbf{mAP} \\
        \specialrule{.1em}{.06em}{.06em}
        I3D \cite{gu2018ava}                      & \xmark  & V+F   &   & K400  & 15.6    \\ 
        ACRN, S3D \cite{sun2018actor}             & \xmark  & V+F   &   & K400  & 17.4    \\ 
        STEP, I3D \cite{yang2019step}             & \xmark  & V+F   &   & K400  & 18.6    \\
        RTPR \cite{li2018recurrent}               & \xmark  & V+F   &   & ImageNet    & 22.3   \\
        Action Transformer, I3D \cite{girdhar2019video} & \xmark& V &   & K400  & 25.0   \\
        LFB, R101+NL \cite{wu2019long}            & \xmark  & V     & v2.1  & K400    & \textbf{27.4}   \\
        SlowFast, R101, 8x8 \cite{feichtenhofer2019slowfast}   & \xmark  & V     &   & K400  & 26.3    \\
        \textbf{YOWO (8-frame)}                   & \checkmark  & V &   & K400  & 15.7     \\
        \textbf{YOWO (16-frame)}                  & \checkmark  & V &   & K400  & 17.2    \\
        \textbf{YOWO (32-frame)}                  & \checkmark  & V &   & K400  & 18.3    \\
        \textbf{YOWO+LFB*}                        & \checkmark  & V &   & K400  & \textbf{19.2}    \\
        \specialrule{.1em}{.06em}{.06em}
        SlowFast, R101+NL, 8x8 \cite{feichtenhofer2019slowfast}  & \xmark      & V &       & K600  & \textbf{29.0}     \\
        \textbf{YOWO (8-frame)}                        & \checkmark  & V &       & K400  & 16.4     \\
        \textbf{YOWO (16-frame)}                       & \checkmark  & V & v2.2  & K400  & 17.9     \\
        \textbf{YOWO (32-frame)}                       & \checkmark  & V &       & K400  & 19.1     \\
        \textbf{YOWO+LFB*}                             & \checkmark  & V &       & K400  & \textbf{20.2}     \\
        \specialrule{.1em}{.06em}{.06em}
    \end{tabular}
    \end{adjustbox}
    \caption{Comparison with state-of-the-art methods on the AVA dataset. Results are reported for frame-mAP under IoU threshold of 0.5. * version of YOWO is non-causal.}
	\label{tab:ava_sota}
\end{table}

\begin{figure}[t!]
	\centering
	 \includegraphics[width = 1.0\linewidth]{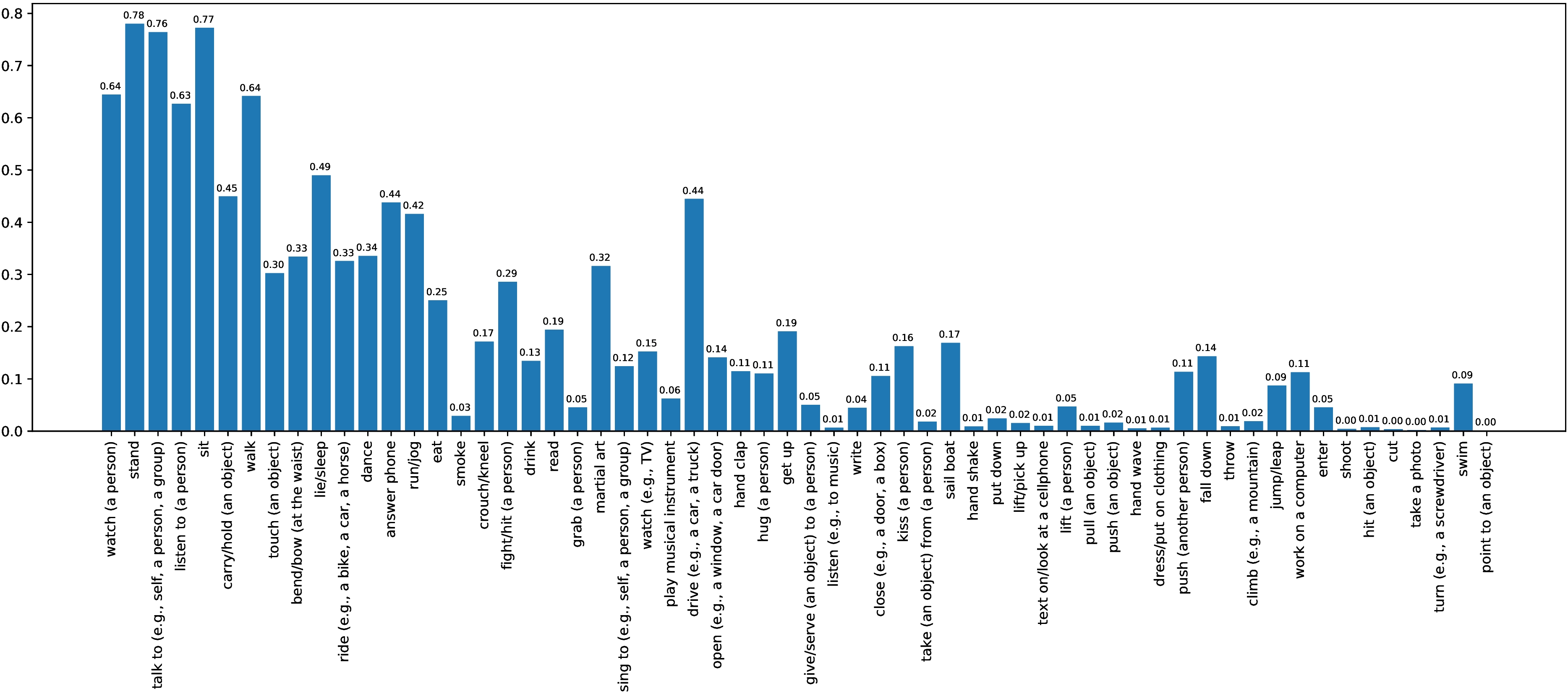}
	\vspace{-1.1cm}
	\caption{Performance of YOWO (32-frames, d=1) on each class of AVA dataset v2.2. Classes are sorted by number of training samples in decreasing order.}
	\label{fig:ava_hist}
\end{figure}

\vspace{0.2cm}
\noindent \textbf{Performance comparison on AVA dataset.} We have compared the performance of YOWO on AVA dataset in Table~\ref{tab:ava_sota}. YOWO is currently the first and only single-stage architecture, which provides competitive results on AVA dataset. All the methods outperforming YOWO are non-causal (i.e. utilizing future frames) and multi-stage architectures mostly utilizing Faster-RCNN architecture. Moreover, these methods either requires high resolution input such as 600 pixels for \cite{li2018recurrent} and 400 pixels for \cite{girdhar2019video}, or strong and computationally heavy SlowFast architecture as 3D-CNN backbone such as \cite{feichtenhofer2019slowfast}. On the other hand, YOWO operates only on the current and previous frames (i.e. causal) with input resolution of $224 \times 224$. Increasing clip size from 8-frames to 32-frames brings an improvement of almost 3\% mAP. Utilization of LFB further improves the performance by around 1\% maP. We also evaluate performance of YOWO (32-frames, d=1) per each class in Fig.~\ref{fig:ava_hist}. The classes are sorted by number of training samples. Although we observe some correlation with the amount of training data, there exist some classes with enough data with poor performance such as smoking.

\vspace{0.2cm}
\noindent \textbf{Runtime comparison} Most of the state-of-the-art methods are two stage architectures, which are computationally expensive to run in real time. YOWO is a unified architecture, which can be trained end-to-end. In addition, we do not employ optical flow, which is computationally burdensome. In Table~\ref{tab:runtime}, we compare runtime performance of YOWO  with other state-of-the-art methods. YOWO's speed is calculated in terms of frames per second (fps) on a single NVIDIA Titan Xp GPU with a batch size of 8. It must be noted that YOWO's 2D and 3D backbones can be replaced with any arbitrary CNN model according to desired runtime performance. Moreover, additional new backbones can be easily introduced for different information source such as \textit{depth} or \textit{infrared} modalities. The only thing to do is modification of CFAM block in order to accommodate new features.

\begin{table}[t!]
    \centering
    \begin{tabular}{lccc}
        \specialrule{.1em}{.2em}{.2em}
        \phantom{aaaaaa} \textbf{Model} & \phantom{a}\textbf{Speed (fps)}\phantom{a} & \phantom{a}\textbf{F-mAP}\phantom{a} & \phantom{a}\textbf{V-mAP}\phantom{a} \\
        \specialrule{.1em}{.2em}{.2em}
        Saha \textit{et al.} \cite{saha2016deep}     & 4                  & -   & 36.4   \\
        ROAD (A) \cite{singh2017online}              & 40                 & -   & 40.9   \\
        ROAD (A+RTF)\cite{singh2017online}           & 28                 & -   & 41.9   \\
        ROAD (A+AF)\cite{singh2017online}            & 7                  & -   & 46.3   \\
        YOWO (8-frames, d=1) & \textbf{62} & 79.2 & 47.6   \\
        YOWO (16-frames, d=1) & 34 & \textbf{80.4} & \textbf{48.8} \\
        \specialrule{.1em}{.2em}{.2em}
    \end{tabular}
    \caption{Run time and performance comparison on dataset UCF101-24 for F-mAP and V-mAP at 0.5 IoU threshold. For YOWO, ResNeXt-101 is used in its 3D backbone.}
	\label{tab:runtime}
\end{table}

\subsection{Model visualization}
\begin{figure*}[t!]
    \centering
    \includegraphics[height=2.20cm,keepaspectratio]{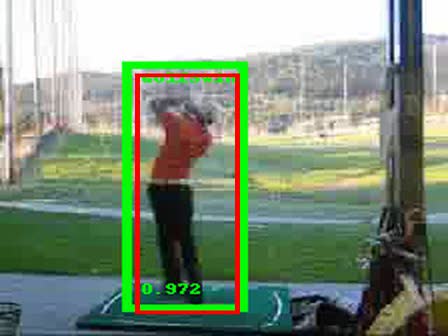}
    \includegraphics[height=2.20cm,keepaspectratio]{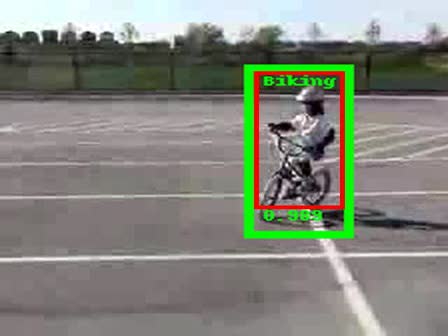}
    \includegraphics[height=2.20cm,keepaspectratio]{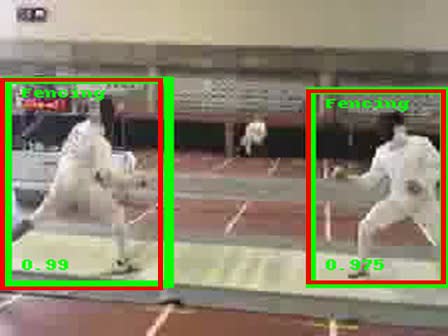}
    \includegraphics[height=2.20cm,keepaspectratio]{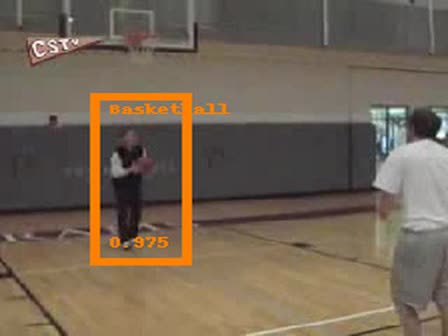}
    \includegraphics[height=2.20cm,keepaspectratio]{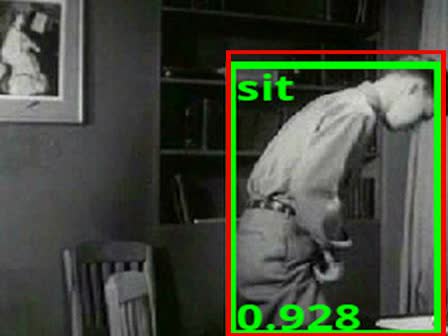}
    \includegraphics[height=2.20cm,keepaspectratio]{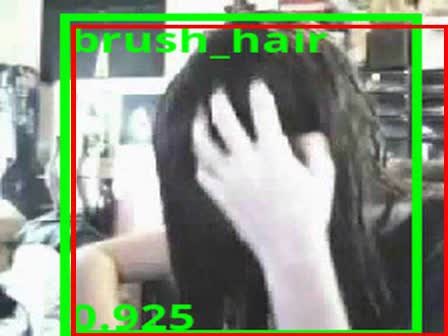}
    \includegraphics[height=2.20cm,keepaspectratio]{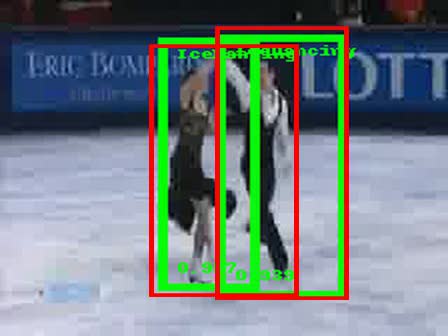}
    \includegraphics[height=2.20cm,keepaspectratio]{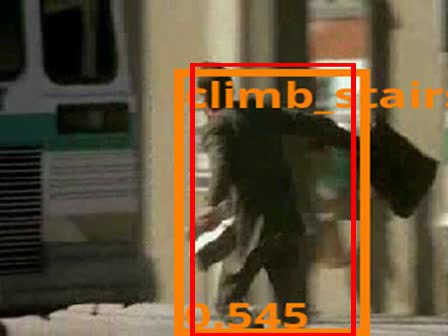}
    \caption{Visualization of action localizations for UCF101-24 and J-HMDB-21 datasets. Red bounding boxes are ground truth while green and orange are true and false positive localizations, respectively.}
    \label{fig:examples}
\end{figure*}

\begin{figure*}[t!]
    \centering
    \includegraphics[height=2.90cm,keepaspectratio]{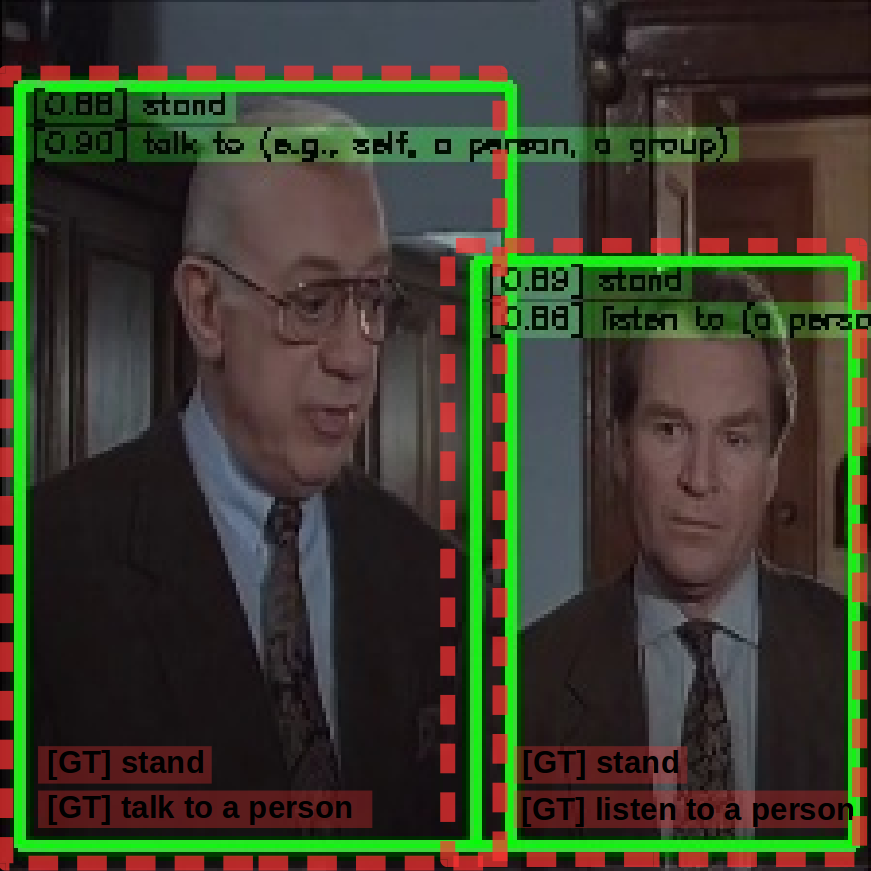}
    \includegraphics[height=2.90cm,keepaspectratio]{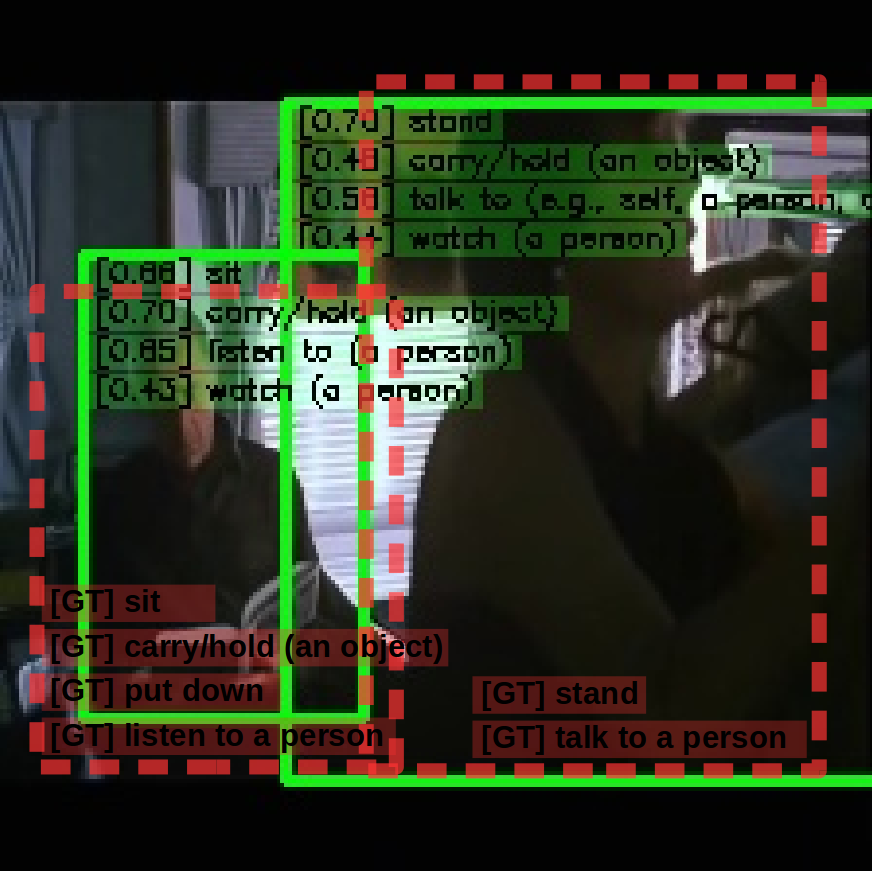}
    \includegraphics[height=2.90cm,keepaspectratio]{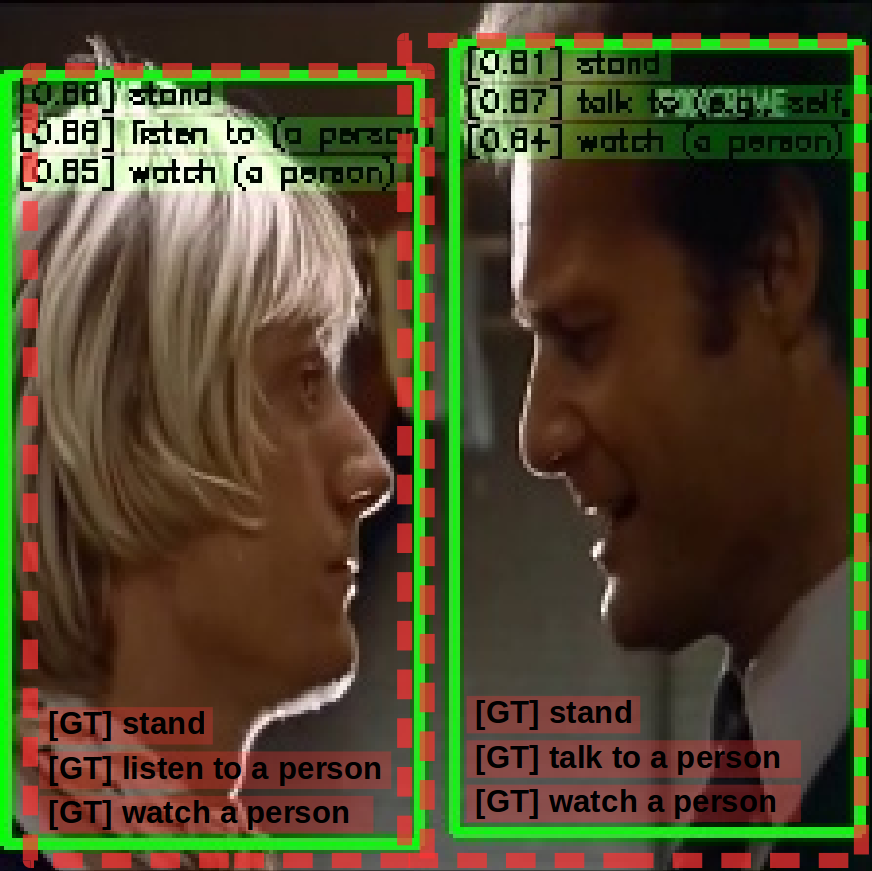}
    \includegraphics[height=2.90cm,keepaspectratio]{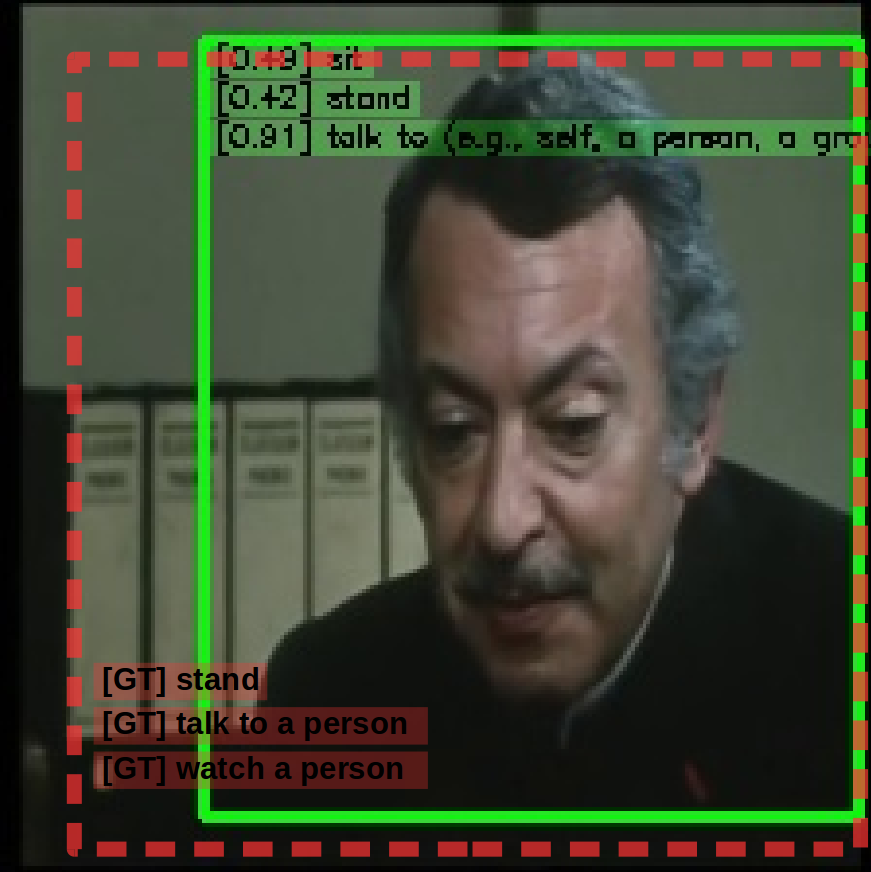}
    \caption{Visualization of action localizations for AVA dataset. Red dashed bounding boxes are ground truth while green bounding boxes are YOWO predictions.}
    \label{fig:ava_examples}
\end{figure*}

In general, YOWO architecture performs a decent job at localizing actions in videos, which is illustrated in Fig.~\ref{fig:examples} for UCF101-24 and J-HMDB-21 dataset; and in Fig.~\ref{fig:ava_examples} for AVA dataset. However, YOWO also has some drawbacks. Firstly, since YOWO produces its predictions according to all the information available at the key frame and the clip, it sometimes makes some false positive detections before the actions are performed. For example, in Fig.~\ref{fig:examples} first row last image, YOWO sees a person holding a ball at a basketball court and detects him very confidently although he is not shooting the ball yet. Secondly, YOWO needs enough temporal content to make correct action localization. If a person starts performing an action suddenly, localization at initial frames lacks temporal content and false actions are recognized consequently, as in Fig.~\ref{fig:examples} second row last image (climbing stair instead of running). Similarly, in the right-most image in Fig.~\ref{fig:ava_examples}, processed clip and key frame does not contain pose information of the person, hence YOWO cannot confidently deduce if the person is sitting or standing. Results in Table~\ref{tab:temporal} confirms that, increasing clip length increases the available temporal information and consequently increases YOWO's performance. LFB is also leveraged for the purpose of increasing temporal content.

\section{Conclusion}
In this paper, we presented a novel unified architecture for spatiotemporal action localization in video streams. Our approach, YOWO, models the spatiotemporal context from successive frames for action understanding while extracting the fine spatial information from key frame to address the localization task in parallel. In addition, we make use of a channel fusion and attention mechanism for effective aggregation of these two kinds of information. Since we do not separate human detection and action classification procedures, the whole network can be optimized by a joint loss in an end-to-end framework. We have carried out a series of comparative evaluations on three challenging datasets, UCF101-24, J-HMDB-21 and AVA, each having different characteristics. Our approach outperforms the other state-of-the-art results on UCF101-24 and J-HMDB-21 datasets while achieving competitive results on AVA dataset. Moreover, YOWO is a causal architecture and can be operated in real-time, which makes it possible to deploy YOWO on mobile devices.

\vspace{-0.2cm}
\section*{Acknowledgements}
We gratefully acknowledge the support by the Deutsche Forschungsgemeinschaft (DFG) under Grant No. RI 658/25-2. We also acknowledge the support of NVIDIA Corporation with the donation of the Titan Xp GPU used for this research.

\bibliography{mybibfile}

\end{document}